\documentclass[nohyperref]{article}

\usepackage{microtype}
\usepackage{graphicx}
\usepackage{subfigure}
\usepackage{dblfloatfix}
\usepackage{booktabs} 

\usepackage{caption}
\usepackage{hyperref}

\usepackage[accepted]{icml2022}

\usepackage{amsmath}
\usepackage{amssymb}
\usepackage{mathtools}
\usepackage{amsthm}

\usepackage[capitalize,noabbrev]{cleveref}

\theoremstyle{plain}

\theoremstyle{definition}

\theoremstyle{remark}

\DeclareMathOperator*{\argmax}{arg\,max}

\usepackage[textsize=tiny]{todonotes}

\icmltitlerunning{Graph neural networks go forward-forward}

\begin{document}

\twocolumn[
\icmltitle{Graph Neural Networks Go Forward-Forward}



\icmlsetsymbol{equal}{*}

\begin{icmlauthorlist}
\icmlauthor{Daniele Paliotta}{yyy}
\icmlauthor{Mathieu Alain}{xxx}
\icmlauthor{Bálint Máté}{yyy}
\icmlauthor{François Fleuret}{yyy}

\end{icmlauthorlist}

\icmlaffiliation{yyy}{Machine Learning Group, University of Geneva}
\icmlaffiliation{xxx}{UCL Centre for Artificial Intelligence }

\icmlcorrespondingauthor{Daniele Paliotta}{daniele.paliotta@unige.ch}
\icmlcorrespondingauthor{Mathieu Alain}{mathieu.alain.21@ucl.ac.uk}
\icmlcorrespondingauthor{Bálint Máté}{balint.mate@unige.ch}


\vskip 0.3in
]


 
\printAffiliationsAndNotice{\icmlEqualContribution} 

\begin{abstract}
We present the Graph Forward-Forward (GFF) algorithm, an extension of the Forward-Forward procedure to graphs, able to handle features distributed over a graph's nodes. This allows training graph neural networks with forward passes only, without backpropagation. Our method is agnostic to the message-passing scheme, and provides a more biologically plausible learning scheme than backpropagation, while also carrying computational advantages. With GFF, graph neural networks are trained greedily layer by layer, using both positive and negative samples. We run experiments on 11 standard graph property prediction tasks, showing how GFF provides an effective alternative to backpropagation for training graph neural networks. This shows in particular that this procedure is remarkably efficient in spite of combining the per-layer training with the locality of the processing in a GNN.

\end{abstract}


\section{Introduction}

Backpropagation (BP, \citealp{Rumelhart1986}) is the \textit{de facto} standard algorithm for training neural networks and has been central to the success of deep learning. Despite its undeniable success, BP does have some drawbacks. For example, the BP algorithm uses the chain rule to compute gradients, which means that non-differentiable components cannot be included in neural networks, and its memory footprint is proportional to the total number of parameters in the model. Moreover, insufficient evidence to support that the BP algorithm could be a biological operation has motivated the investigation of \textit{neuromorphic} methods for updating the parameters of neural networks \cite{spiking, Whittington2019, Dellaferrera2022}. The neuromorphic approach in deep learning is a growing trend that aims to relate deep learning methods to brain processes. The intent is to mimic the functioning of the human brain and the neurons that compose it. A major motivation behind the recent neurophormic development is the observation that the brain is highly capable and yet needs low energy compared to the computers needed to run and use large neural networks.

Recent work by \citet{Hinton2022} challenges the dominance of the BP algorithm by proposing the forward-forward (FF) algorithm as an alternative for training neural networks. The FF algorithm is inspired by Boltzmann machines \cite{Hinton1986} and by contrastive learning methods such as \citet{pmlr-v9-gutmann10a} and constitutes a better candidate in terms of biological plausibility. Initial experiments show that the FF algorithm achieves strong performance on vision datasets like MNIST and CIFAR-10, and that it can be extended to dealing with sequences \cite{Hinton2022}.


Instead of a forward and a backward pass, the FF algorithm performs a pair of forward passes to update the parameters of a network. The two passes run on two different datasets and have opposite objectives. The positive (negative) pass uses a positive (negative) dataset and adjusts, greedily layer by layer, the parameters of the neural network by maximizing (minimizing) a value, called the \textit{goodness}. The FF algorithm is able to incorporate non-differentiable components between the layers and intends one day to enable the development of large power-efficient neural networks. The FF algorithm is able to handle both the supervised and unsupervised learning paradigms. This promising algorithm, however, remains largely unexplored at the moment. 

Our main contribution is to improve and extend the FF algorithm to graph neural networks. We develop a recipe to calculate the \textit{goodness} of any attributed graph and introduce the forward-forward algorithm on graph neural network (GFF). In the setting of graph property prediction, we show that GFF offers comparable performance to standard backpropagation on a range of different datasets from various domains. Additionally, we analyze different methods of computing the graph goodness, and of incorporating ground truth information in the message-passing framework. Finally, we provide statistics and discussions on the computational advantages of GFF.


\section{Related Work}
\subsection{Graph Neural Networks} A graph $G$ is a pair $(V,E)$ where $V =\{v_1, ... ,v_n\}$ is a set of nodes and $E$ is a set of pairs of nodes $(v_i,v_j)$ called edges.
Graphs are ubiquitous in the real world and can be used to represent a large number of entities such as molecules, social networks, and maps. The abundance of problems related to graphs in numerous areas, including the natural and social sciences, has fuelled the development of methods to leverage the properties of graphs. Graph neural networks (GNNs, \citealp{Scarselli2009}) have quickly become the standard for dealing with graph-structured data.  
This large family of models has already generated several impressive achievements \cite{DP2021, Davies2021, Zhou2020b, surveygnn}. GNNs operate through an iterative procedure called \textit{message-passing} (MP). The representation of each node is updated, in parallel, by computing a function that aggregates the neighboring nodes. This procedure is repeated multiple times, where each iteration is parametrized by a layer in the GNN. The more layers there are, the more informed the remote nodes are of each other. After several message-passing iterations, a pooling function can be applied to aggregate the node feature vectors, with the resulting vector being used to make the final prediction.

\subsection{The Forward-Forward algorithm}

\paragraph{Positive and negative datasets.}
The FF algorithms requires encoding the label information in the data points. In \citet{Hinton2022}, the one-hot encoded label is inserted in a corner of the image that contains no information. The correct label encoding is added to an image to form a positive sample. An incorrect label encoding is added to an image to form a negative sample. There is, of course, only one possible positive dataset. There can, however, be more than one negative dataset.

\paragraph{Goodness Function.} One of the building blocks of the FF algorithm is the goodness function. This function is needed to update the parameters of the neural network. In \citet{Hinton2022}, the goodness of a layer is the sum of the squared activities. The advantage of this definition is that it is simple and requires little computation. For example, the goodness function using squared activations is more efficient than the free energy function in the Boltzmann machines and has simple derivatives. 

\subsection{Contrastive Learning}
The idea of contrastive learning is to learn by contrasting data that are similar and dissimilar \cite{Chen2020}. This is often achieved by augmenting a dataset to create a similar, but slightly different, dataset. In computer vision, this augmentation can be done by transforming the images (cropping, rotating, and recoloring). This principle has been successfully adapted to GNNs in instances such as GraphCL \cite{You2020} and MolCLR \cite{Wang2022}. These methods require finding a method to compare two representations (vectors). In the FF algorithm, contrastive learning is applied by using positive and negative datasets described above. The advantage of FF-based algorithms is that it is not required to compare the (vector) representations, but only the (scalar) goodness values.


\section{Method} \label{sec:Methods}

\begin{figure}
\centering
\includegraphics[scale=0.70]{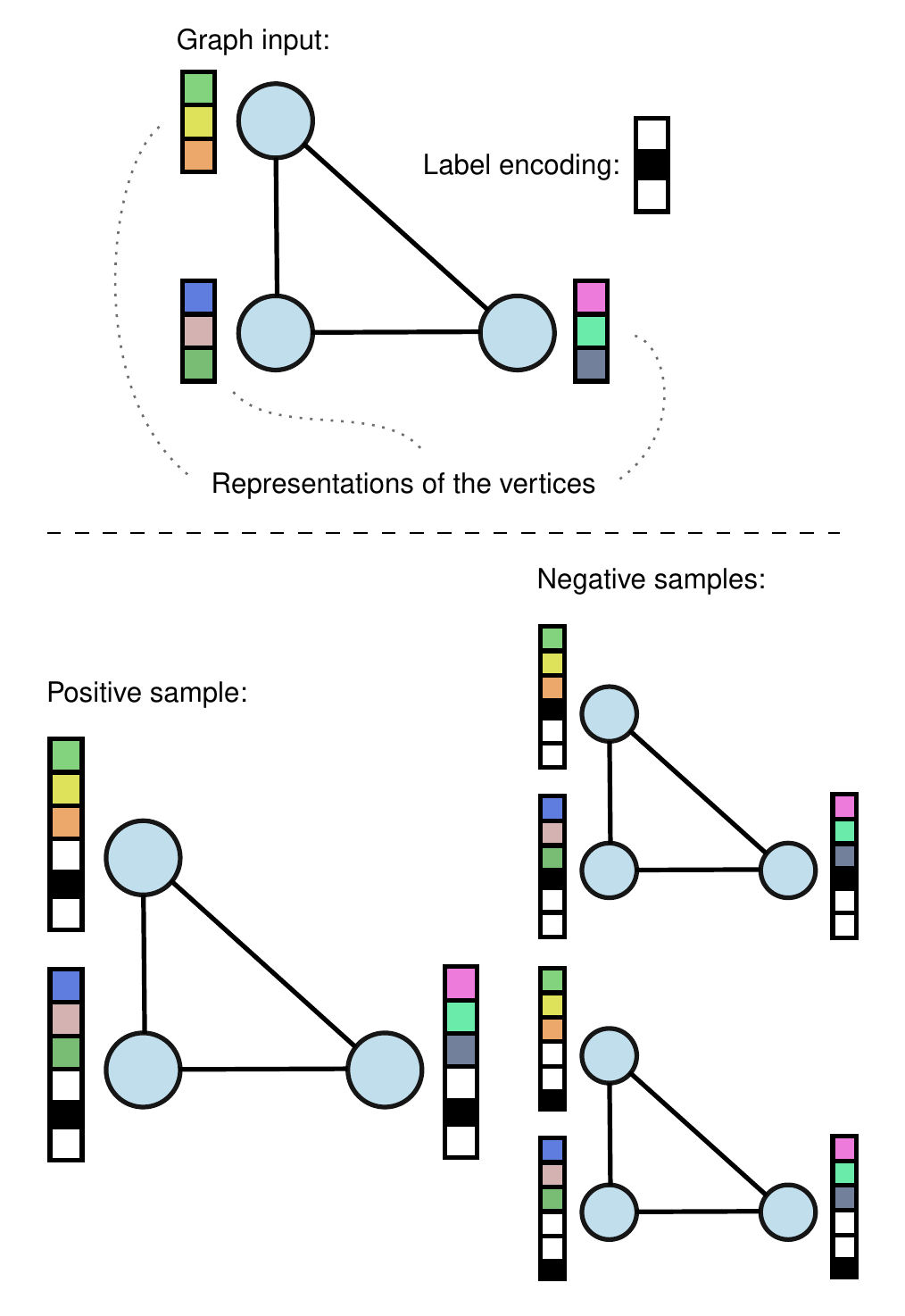}
\caption{On the top: A graph input and the associated label encoding. The graph is composed of three nodes each having a representation vector. The same sample is transformed into a positive sample and two negative samples. On the bottom-left (positive sample): the correct label encoding has been added to the representation vectors of the nodes. On the bottom-right (negative sample): the incorrect label encoding has been added to the representation vectors of the nodes.}
\label{fig:pndatasets}
\end{figure}

This section describes the GFF recipe in detail. Our work focuses on the task of graph property prediction. The initial datasets are composed of multiple graphs that each belong to a label.

\begin{figure}
\centering
\includegraphics[scale=0.78]{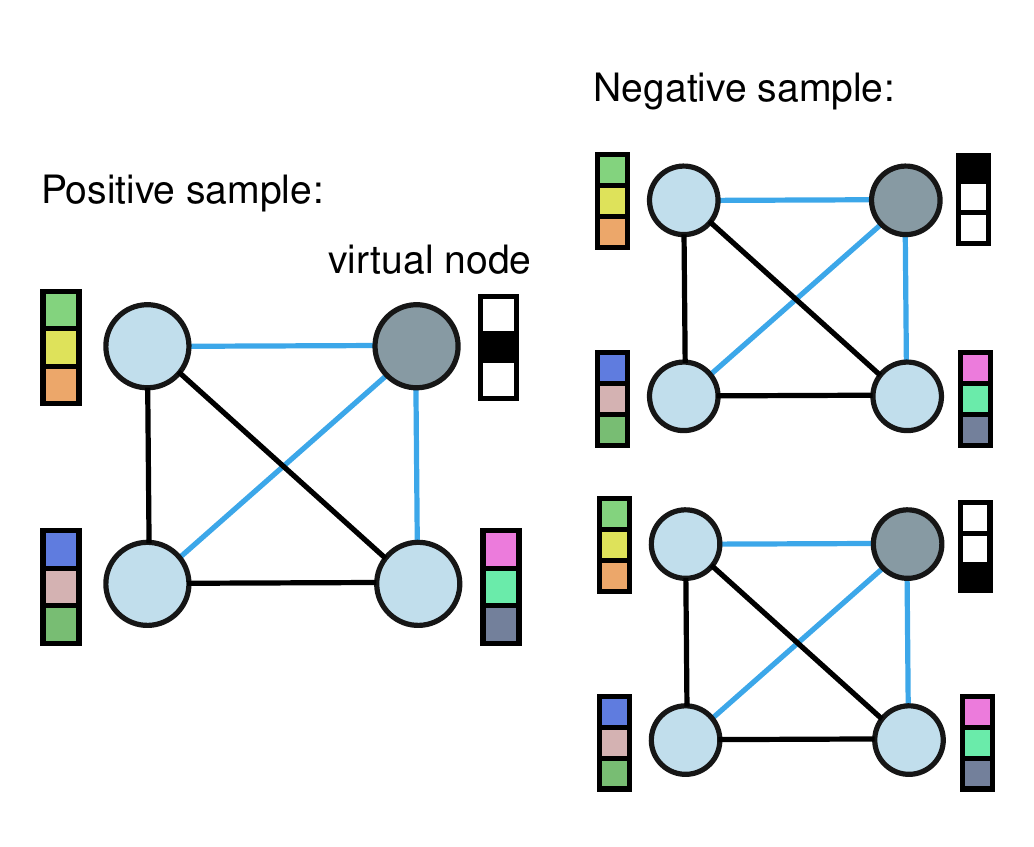}
\caption{The same sample found in Figure \ref{fig:pndatasets}, but transformed using a virtual node into a positive sample and two negative samples. On the left (positive sample): the correct label encoding has been added to the representation vectors of the nodes. On the right (negative sample): the incorrect label encoding has been added to the representation vectors of the nodes.}
\label{fig:virtual}
\end{figure}

In the standard FF algorithm, the different labels are represented by a one-hot encoding scheme. In \citet{Hinton2022} and subsequent works, the label encodings are inserted as replacements for the first few pixels (which contain no information, of course) of the input images. In order to include label information in a graph, we develop and test two approaches. One consists in appending the one-hot encoded label to each node representation in the input graphs. This encoding scheme is displayed in Figure \ref{fig:pndatasets}.
A second approach, which is more graph-specific and which requires no data redundancy, is to endow input graphs with an additional \textit{virtual node} carrying all information about the label. A virtual node is a node added to the graph and connected in some ways to the rest of the nodes. This encoding scheme is described in Figure \ref{fig:virtual}. Note that the virtual node is connected bidirectionally to every other node in the graph. In order to allow the network to treat the virtual node as a separate entity, we use separate weight matrices for the message passing along the virtual node edges.

In order to generate negative samples for training with GFF, we simply encode the input graph with an incorrect label. Since the only difference between positive and negative data is the label, GFF should learn to ignore all features of the graph that do not correlate with the label.

\paragraph{Notation.} 

Let $n$ be the number of nodes in the input graph and suppose that this number remains the same throughout the layers of the network. Let  $d$ be the dimension of the activation vectors. Let $\mathbf{A}^{(i)} = \left(\mathbf{a}^{(i)}_1, \dots, \mathbf{a}^{(i)}_n\right)$ be the activation matrix that contains the activation vectors $\mathbf{a}^{(i)}_j= \left(a^{(i)}_{j1}, \dots, a^{(i)}_{jd}\right)$ of all the nodes of the $i$-th layer. 

\subsection{Goodness of a Graph} Learning on graphs using FF requires defining the goodness function for an attributed graph. This is the function that allows computing the layer-wise loss. Intuitively, the network is trained so that the goodness for positive samples is high (or higher than a threshold), while the goodness for negative samples is minimized. Moreover, the goodness should be computed in such a way that all information about it is discarded before passing the activations to the next layer. More information about how the goodness is used for training and inference is provided in Sections \ref{training} and \ref{inference}.
One of the main differences with the approach proposed in \citet{Hinton2022} is that, when dealing with graphs, we carry a separate feature vector for each node, and not a single vector representing the whole sample. Since the goodness comes in the form of a single scalar for the whole graph, we need to find a way to aggregate these vectors. The aggregation function can be any permutation-invariant function of the set of node representations. Then, the mean of the elements of the resulting vector represents the goodness. 

The goodness $g^{(i)}$ of the $i$-th layer, when we use sum-pooling as an aggregation function, is defined such that
\begin{equation}
    \label{eq:goodness}
    g^{(i)} = \frac{1}{d} \sum_{k=1}^{d} \sum_{j=1}^{n} \left(a^{(i)}_{jk}\right)^2.
\end{equation}
Notice that more than one goodness function is possible, but the one above shines by virtue of being uncomplicated, and we find that it works well in practice. 
To summarize, the computation of the goodness is based on the following recipe:
\begin{enumerate}
    \item Normalized activity vectors of all the nodes and the adjacency matrix are fed to the message-passing algorithm. This produces updated activity vectors. A nonlinear activation function is applied.
    \item The activity vectors are squared.
    \item The squared activity vectors are aggregated with a pooling function. This produces an aggregated vector.
    \item The mean of the aggregated activity vector is computed and represents the goodness of the graph.
\end{enumerate}

\subsection{Architecture}


\begin{figure}
\centering
\includegraphics[scale=0.63]{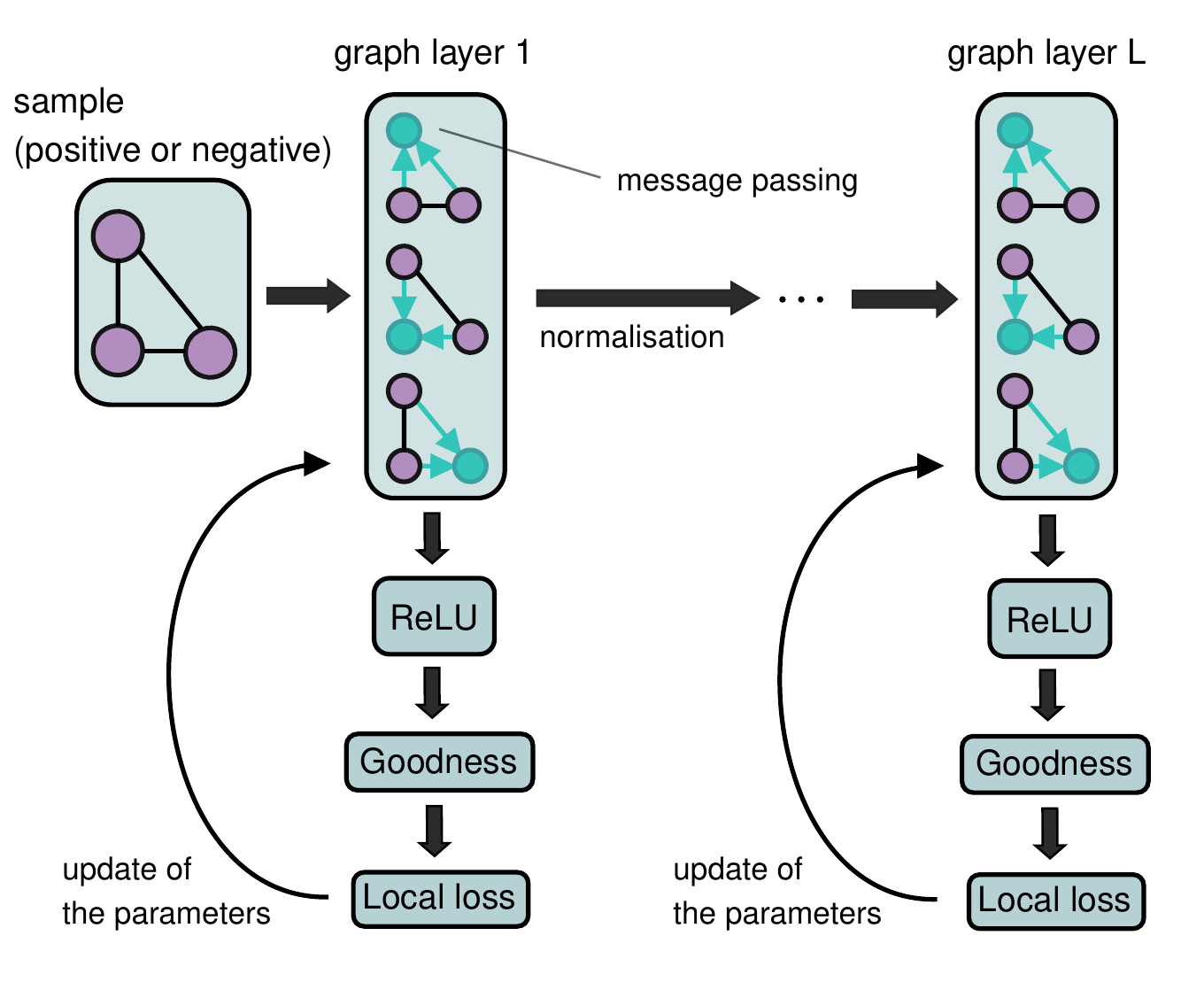}
\caption{The architecture of an FFGNN. The layers are updated in a greedy manner. Each layer performs the MP procedure on each node (illustrated by the multiple graphs in each layer).}
\label{fig:ffgnn}
\end{figure}

Our framework can potentially be applied to any GNN. The layers can be convolutional or even attentional. In our experiments, the layers employed are convolutional layers found in the popular graph convolutional networks (GCN, \citealp{Kipf2017}). The main feat of GCNs is to extend the convolution operator to non-Euclidean data in the form of graphs. In GCNs, for every node in the input graph, neighbor representations are aggregated with a position-invariant function. A linear transformation is applied to the resulting, aggregated vector, and the nodes are then updated with the new contextual representation. Specifically, the node-level update is described by the following equation:
\begin{equation}
 \mathbf{x}^{\prime}_i = \mathbf{\Theta}^{\top} \sum_{j \in
        \mathcal{N}(v) \cup \{ i \}} \frac{ \mathbf{x}_j}{c_{i,j}},
\end{equation}
where $\mathbf{x}_j$ is the vector representation of node $i$, $\mathcal{N}(i)$ is the set of neighbors of node $i$, $\mathbf{\Theta}$ is the weight matrix for the layer, $c_{i,j}$ is a normalization constant for the edges between node $i$ and $j$, and $\mathbf{x}^{\prime}_i$ is the updated node representation. 

In the GFF algorithm, the activity vectors are normalized before being passed to the next layer. This is performed in order to avoid the information used to compute the goodness in the previous layer influencing the goodness in the next one. In this paper, the norm used is the length of the vector. Remember that $\mathbf{a}^{(i)}_j =\left(a^{(i)}_{j1}, a^{(i)}_{j2}, \dots, a^{(i)}_{jd}\right)$ is the activity vector of the $j$-th node of the $i$-th layer. The norm is defined such that  
\begin{equation}
    \left\|\mathbf{a}^{(i)}_j\right\| = \sqrt{\left(a^{(i)}_{j1}\right)^2 + \left(a^{(i)}_{j2}\right)^2 + \cdots + \left(a^{(i)}_{jd}\right)^2}.
\end{equation}
This norm has the advantage of being simple, but one could of course try other types of layer normalization. 

\subsection{Loss and Training} \label{training}

One of the main differences between a GFF-based GNN and a standard GNN is that the parameters of the former are updated layer by layer, instead of after going through all the layers. In other words, only one layer is being trained at each time. In this sense, the GFF algorithm can be said to be greedy or local layer-wise. Note that there is no predictive loss function. The objective is to maintain the goodness above a threshold if the data is positive and below this threshold if the data is negative. The idea is to minimize or maximize (depending if positive or negative) the local loss function of the layer. Let $T$ be a threshold. For positive samples, the local loss function of the $i$-th layer is defined such that
\begin{equation}
     L_i = \log \left(1+\exp{\left(-g^{(i)} + T\right)}\right).
\end{equation}
For negative samples,
\begin{equation}
     L_i = \log \left(1+\exp{\left(g^{(i)} - T\right)}\right).
\end{equation}
The update can then be directed by the derivative of the local loss function. Note that this requires computing derivatives, but not applying the chain rule through all the layers. This is the reason that non-differentiable components can be added between the layers. This could allow, for example, symbolic components \cite{Garcez2015, Lamb2020}. Once the parameters of a layer are totally updated, it is the turn of the next layer to follow the same procedure. The activity vectors of the previous layer are passed to the next layer. The entire architecture can be visualized in Figure \ref{fig:ffgnn}. 

This greedy approach using a local loss function should result in the GFF algorithm being more memory efficient during training since we need to compute much fewer gradients and we don't need to keep the activations in memory for the backward pass. We elaborate more on this in Section \ref{sec:Discussion}. The optimization problem also gets simpler, given that  each layer is solving a smaller task in a lower-dimensional parameter space. 

\subsection{Inference} \label{inference}

The prediction phase in GFF is done in a quite different manner than in backprop-GNNs since there is no predictive loss function. Let us suppose that the goal is to classify some graph $G$. Let $C$ be the number of possible labels. The idea is to first create the \textit{label graph} $G_{c_i}$ for each label $c_i$. The label graph $G_{c_i}$ is created by appending the label encoding of the label $c_i$ to all the nodes of the graph $G$. Let $D_C$ be the augmented dataset containing the $C$ label graphs. The next step is to compute the \textit{label goodness} of each graph in the set $D_C$. The label goodness $g_{c_i}$ of the label graph $G_{c_i}$ is the sum of the goodness of all the layers of the FGNN that received $G_{c_i}$ in input. Let $L$ be the number of layers of the FGNN.
\begin{figure}[h!]
\centering
\includegraphics[scale=0.73]{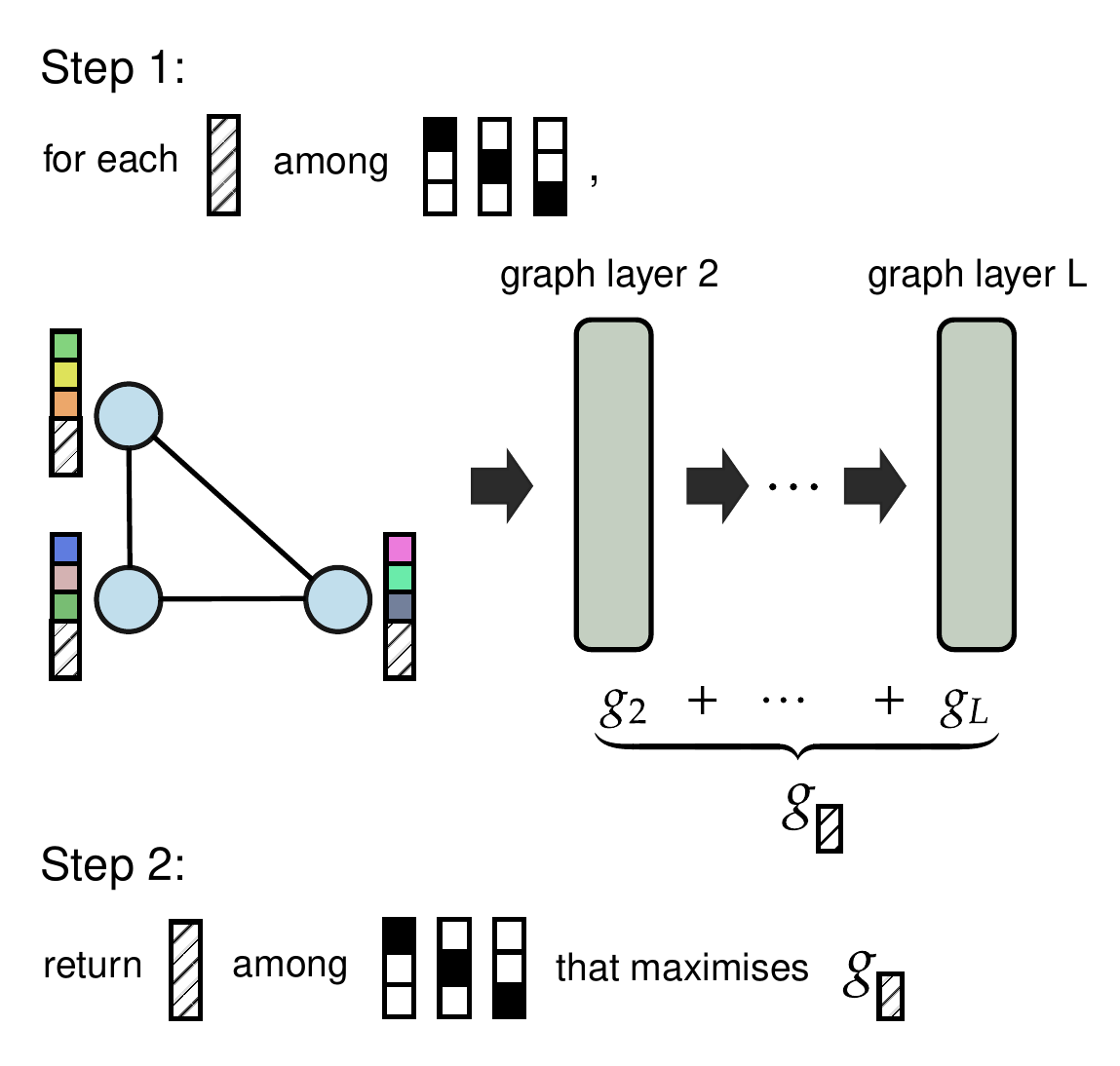}
\caption{The prediction is done in two steps. Step 1: the label goodness is computed for each label. Step 2: the label selected is the label that has the highest label goodness.}
\label{fig:pred}
\end{figure}

More formally, the label goodness $g_{c_i}$ is defined such that
\begin{equation}
     g_{c_i} = \sum^{L}_{i=2} g^{(i)}.
\end{equation}
Note that, similar to \citet{Hinton2022}, the goodness of the first layer is not used in the prediction. This leads to better empirical results. The prediction is then accomplished by computing the label goodness of each of the label graph in the set $D_C$ and then choosing the label that possesses the highest label goodness. More formally, the returned label $c^\star$ is selected such that
\begin{equation}
     c^\star = \argmax_{c_i} g_{c_i}.
\end{equation}
For an illustration of the prediction scheme, see Figure \ref{fig:pred}. This type of prediction is interesting since the goodness of the other labels can also provide information. However, from a computational point of view, it can become difficult if the number of labels is very large. The adaptation of this kind of prediction scheme to a regression setting could be of value and is the subject of future work.

\section{Experiments}\label{sec:Exp}
We run multiple experiments on 11 datasets for graph property prediction. We compare the performances of GNNs trained with backpropagation with GNNs trained using GFF. We then perform several ablations to better understand and evaluate the building blocks of our method such as the pooling function, the goodness computation, and the label encoding.
For datasets where the graph doesn't have node features, we assign the one-hot encoded node degree as the initial feature for each node.

All experiments are implemented in Pytorch/Pytorch Geometric \cite{Fey/Lenssen/2019, NEURIPS2019_9015_pytorch} on 1 NVIDIA 3090 GPU.

\subsection{Datasets}

\begin{table*}[ht]\centering 
\begin{tabular}{ccccccc}
\hline
 \textbf{Dataset} & \textbf{\# of graphs} & \textbf{\# of labels} & \textbf{avg \# of nodes} & \textbf{avg \# of edges} & \textbf{num feat. (node)} & \textbf{avg degree} \\
\hline
 PROTEINS & 1113 & 2 & 39.06 & 72.82 & 4 & 7 \\
 IMDB-Binary & 1000 & 2 & 19.77 & 96.53 & 501 & 8 \\
 BZR & 405 & 2 & 35.75 & 38.36 & 53 & 4  \\
 COX2 & 467 & 2 & 41.22	 & 43.45 & 53 & 4 \\
 MUTAG & 188 & 2 & 17.93 & 19.79 & 7 & 4 \\
 SN12C & 40004 & 2 & 26.08 & 28.11 & 65 & -- \\
 SW-620H & 40532 & 2 & 46.62 & 48.65 & 66 & -- \\
 Yeast & 79601 & 2 & 21.64 & 22.84 & 74 & 2 \\
 Peking\_1 & 85 & 2 & 39.31 & 77.35 & 190 & 7 \\
 COLLAB & 5000 & 3 & 74.49 & 2457.78 & 501 & 263 \\
 MSRC\_9 & 221 & 8 & 40.58 & 97.94 & 10 & 9 \\

\hline
\end{tabular}
\caption{Information and summary statistics for the various datasets used in this paper.}
\label{table:datasetsinfo}
\end{table*}

The datasets used in this paper are mainly obtained from the TUDatasets collection \cite{TUDatasets}. The datasets range from relatively small ones such as \textbf{Peking\_1} to larger ones such as \textbf{Yeast} and cover molecules (\textbf{BZR}, \textbf{COX2}, \textbf{MUTAG}, \textbf{SN12C}, \textbf{SW-620H}, \textbf{Yeast}), bioinformatics (\textbf{PROTEINS}, \textbf{Peking\_1}), computer vision (\textbf{MSRC\_9}) and social networks (\textbf{COLLAB}, \textbf{IMDB-Binary}). For more information on the datasets, see Table \ref{table:datasetsinfo}.

\subsection{Architecture and Training Details}
In our experiments, all models contain 3 GCN layers with 128 hidden units per layer. We train the models for 200 epochs with a batch size of 128, using the Adam  optimizer \cite{KingmaB14_adam} with learning rate $10^{-3}$. We report the mean and the standard deviation over 5 different random seeds.

\subsection{Results}

\begin{table}[h!]\centering\small
\begin{tabular}{@{}lcccc@{}}\toprule
\textbf{Dataset}  & \textbf{GCN} & \textbf{FF-GCN} \\
\midrule
PROTEINS & $0.60 \pm 0.02$ & $0.62 \pm 0.10$ \\
IMDB-Binary & $0.70 \pm 0.05$ & $0.63 \pm 0.08$ \\
BZR  & $0.92 \pm 0.03$ & $0.89 \pm 0.05$\\
COX2  & $0.82 \pm 0.04$ & $0.78 \pm 0.08$\\
MUTAG  & $0.71\pm  0.04$ & $0.71\pm  0.04$\\
SN12C  &$0.96\pm 0.00$ & $0.95 \pm 0.00$ \\
SW-620H  &$0.95\pm 0.00$ & $0.94 \pm 0.00$ \\
Yeast &$0.88\pm 0.00$ & $0.88 \pm 0.00$  \\
Peking\_1   &$0.67 \pm 0.11$ & $0.59 \pm 0.06$\\
COLLAB   & $0.79 \pm 0.01$ & $0.68 \pm 0.02$\\
MSRC\_9  &$0.91 \pm 0.04$ & $0.82 \pm 0.09$\\
\bottomrule
\end{tabular}
\caption{Comparison of the GCN (BP) and the FF-GCN (GFF). The results are in terms of test accuracy and have been obtained on 5 random seeds.}
\label{table:results}
\end{table}

Table \ref{table:results} shows the comparison between GCNs trained with backpropagation and a GCNs adapted and trained with GFF. We find that the two networks perform very similarly in almost all cases. In one instance, we find that GFF outperforms the backpropagation alternative. Not that in this table for GFF we only show the use of sum-pooling to compute the goodness (see Section \ref{sec:Methods}). However, for some datasets, we get even better results with mean pooling, which we show in the ablations (Section \ref{Ablation}).

\begin{table}[h!]\centering\small
\begin{tabular}{lc}\toprule
\textbf{Dataset} & \textbf{Speedup} \\
\midrule
PROTEINS & 22 \\
IMDB-Binary & 18 \\
BZR  & 35 \\
COX2 & 22 \\
MUTAG &  61\\
SN12C & 2.5 \\
SW-620H & 2.7 \\
Yeast & 2.6 \\
Peking\_1 & 42 \\
COLLAB  & 2.9 \\
MSRC\_9 & 28 \\
\bottomrule
\end{tabular}
\caption{Comparison of training speed between backpropation and FF. The speedup value represents the factor by  which the GFF algorithm is faster to train than the GCN for the same number of epochs.}
\label{table:speedup}
\end{table}

In Table \ref{table:speedup}, we report the difference in training speed between training with backpropagation and with FF. Note that we only compare the amount of time it takes to perform the same number of epochs on the datasets, and not the time it takes for the models to converge. For small datasets we observe speedups of more than an order of magnitude. For larger datasets, the time spent moving data in and out of the memory takes a significant portion of the training time, and the relative speedup of FF compared to backpropagation decreases. Nonetheless, on all datasets, FF requires at most half as long to perform the same number of training steps as backpropgation.

\subsection{Ablations} 
\label{Ablation}

\paragraph{Virtual node.} As explained in Section \ref{sec:Methods}, we test another method for encoding the label in the samples, based on virtual nodes. Each input graph is endowed with an virtual 
node that is connected bidirectionally to every other node. The virtual node features are initialized with a fixed embedding of the ground truth label for the graph. Additionally, we need the network to treat the virtual node as a separate entity, and thus learn to propagate its message separately. To this purpose, we use a Relational GCN layer (RGCN, \citealp{rgcn}), which has separate weights for different edge categories. In our case, the 
virtual node edges are assigned different categories than standard edges.
Finally, the goodness is computed by only considering the activities of the virtual node after message-passing. Conceptually, the virtual node approach has the advantage that it avoids duplicating the data, while also being more interpretable. Table \ref{table:virtual_ablation} shows the results of the ablation on the \textbf{PROTEINS} and \textbf{COLLAB} datasets. We find that the node feature concatenation method still works better and more consistently. Additionally, we experimentally find that pairing an RGCN with our GFF loss causes the training to be quite unstable and prone to reaching infinite values in the exponential. 

\begin{table}[H]\centering\small
\begin{tabular}{lcc}\toprule
\textbf{Dataset} &  \textbf{Concat} & \textbf{Virtual}\\
\midrule
PROTEINS & $0.62\pm 0.10$  & $0.61\pm 0.05$ \\
COLLAB  & $0.68 \pm 0.02$ & $0.52 \pm 0.01$ \\
\bottomrule
\end{tabular}
\caption{Results for ablation on the label encoding method. Test accuracy is shown. The concatenation method outperforms the virtual node approach.}
\label{table:virtual_ablation}
\end{table}

\paragraph{Computation of the Goodness.} One of the key components of  training with the FF algorithm is the notion of goodness of a (sample, label) pair. At each graph layer, there are multiple reasonable ways of computing the goodness. One can first square the representation component-wise and then sum over both the node and feature dimensions (Eq. \ref{eq:goodness}). Alternatively, one can first sum over the node dimension to build one vector for the whole graph, square it component-wise, and finally sum it,
\begin{equation}
    \label{eq:goodness2}
    \hat{g}^{(i)} = \frac{1}{d} \sum_{k=1}^{d} \left(\sum_{j=1}^{n} a^{(i)}_{jk}\right)^2.
\end{equation}

In our experiments we computed the goodness values using Eq. \ref{eq:goodness}. Table \ref{table:goodness} shows how the choice the goodness computation influences performance on MSRC\_9, IMDB-BINARY, BZR and COX2.
\begin{table}[H]\centering\small
\begin{tabular}{lcc}\toprule
\textbf{Dataset} &  \textbf{square-pool-mean} & \textbf{pool-square-mean}\\
\midrule
MSRC\_9 & $0.90 \pm 0.05$ & $0.80 \pm 0.08$\\
IMDB-BINARY &$0.59 \pm 0.06$ & $0.50 \pm 0.06$ \\
BZR &$0.78\pm 0.02$ & $0.77 \pm 0.10$ \\
COX2  &$0.81\pm 0.05$ & $0.77 \pm 0.09$ \\
\bottomrule
\end{tabular}
\caption{Comparison of the different ways of computing the goodness. Eq. \ref{eq:goodness} corresponds to the column ``square-pool-mean'' and Eq. \ref{eq:goodness2} correspond to the column ``pool-square-mean''.}
\label{table:goodness}
\end{table}

\paragraph{Mean vs. Additive Pooling.} As discussed in the previous paragraph, one way or another a pooling step over the nodes of the graph has to be applied when computing the goodness. In our experiments, we used additive pooling. In this section, we compare how changing the pooling to mean-pooling influences performance.  We didn't observe a statistically significant difference between these two pooling methods, the numerical results are given in Table \ref{table:pooling}.
Note that additional pooling operations that we didn't investigate (for instance max-pooling)  can also be used as alternative ways of computing the goodness.

\begin{table}[H]\centering\small
\begin{tabular}{lcc}\toprule
\textbf{Dataset} &  \textbf{additive} & \textbf{mean}\\
\midrule
MSRC\_9 & $0.90 \pm 0.05$ & $0.87 \pm 0.07$\\
IMDB-BINARY &$0.59 \pm 0.06$ & $0.61 \pm 0.07$ \\
BZR &$0.78\pm 0.02$ & $0.84 \pm 0.07$ \\
COX2  &$0.81\pm 0.05$ & $0.77 \pm 0.06$ \\
\bottomrule
\end{tabular}
\caption{Comparison of additive-pooling and mean-pooling when computing the goodness. Note that mean-pooling and additive pooling are not equivalent as the graphs in the dataset might have a different number of nodes. }
\label{table:pooling}
\end{table}


\section{Discussion} 
\label{sec:Discussion}

\subsection{Memory Footprint}
The memory footprint of training a graph neural network with GFF can be significantly lower that the backpropagation alternative. When using backpropagation, all of the activations have to be kept in memory to allow the propagation of the gradients during the backward pass. This results in memory consumption increasing as we add more layers to the network. For very large networks, this can occupy a very large amount of memory, making it impossible to train the model on a single GPU and requiring complex engineering to train in a distributed way. When training networks with FF, only a single layer is trained at a time, and thus only a subset of the activations need to be kept in memory. For deep networks, the difference in memory requirements can be significant. Provided that GFF scales, it would be possible to train very deep networks on very large graphs, such as social networks or protein complexes, with a fraction of the memory needed by backpropagation.

\subsection{GFF and Top-down Effects}
One weakness of GFF with respect to backpropagation, as outlined in \citet{Hinton2022}, lies in the lack of top-down information flow.  In practice, what is learned in later layers cannot affect what is learned in earlier layers. This is a limitation for GNNs trained with GFF, where long-range dependencies captured in later layers cannot inform the parameter updates of the initial layers. An initial recipe was proposed in \cite{Hinton2022}. Nonetheless, the graph structure of our data might allow augmenting the input with additional connections, that still allow capturing long-range context, despite missing top-down propagation. The investigation of this question is left for future work.

\subsection{Generating Negative Data}
Negative samples are fundamental to the functioning of the FF-based algorithms. As of now, the main method of coming up with negative samples has been to assign the wrong label to existing samples. However, this seems somehow limiting. Drawing from the contrastive learning literature, it might be possible to generate negative samples by modifying the underlying graph. This might for example help GFF focus on the longer-range correlations that have been shown to be vital for some GNN applications.

\section{Conclusion}

We introduced the Graph Forward-Forward algorithm, which extends the Forward-Forward algorithm to graph neural networks. The empirical findings in Section \ref{sec:Exp} demonstrate that  our method is a viable alternative to backpropagation for GNNs. Despite some limitations, GFF carries several advantages as it's more biologically plausible and computationally more efficient in certain settings. Comprehensive analysis of the building blocks of our method further clarifies the inner workings of the models.
This is quite encouraging and we hope and expect that it opens the doors for further research in this direction. 

\section{Future work.} In light of the novelty of the FF algorithm and the fact that this paper is the first one to extend the FF algorithm to graphs, there are, of course, many open problems and exciting possible directions to take in the future. 

\paragraph{Goodness and local loss functions.} A good question is how to choose the goodness function and the local loss function. In this paper, the squared activity vectors are added to each other, but it could be interesting to explore other aggregation methods. There are multiple options and it would not be too surprising if it turned out that the choice of the goodness function and the local loss function depends on the specific application.

\paragraph{Extension to other graph tasks.}
This paper focuses on the supervised learning paradigm and more precisely on the graph property prediction task. The FF algorithm, and by extension the GFF algorithm, could also be used in an unsupervised setting. The unsupervised paradigm, however, raises some challenges in terms of how to create the negative datasets. In the supervised learning paradigm, nevertheless, the method could be extended to handle graph regression. The main issue here is how to define the negative dataset and how to perform the prediction, considering that the number of classes is no longer finite. Node and edge-level tasks are also a natural extension of the method. We leave it for future work.

\section*{Acknowledgements}

The authors acknowledge support from the Swiss National Science Foundation under grant number CRSII5\_193716 - “Robust Deep Density Models for High-Energy Particle Physics and Solar Flare Analysis (RODEM)”

\bibliography{GNNsGoFF}
\bibliographystyle{icml2022}

\newpage
\appendix
\onecolumn

\end{document}